# Collaborative Filtering and the Missing at Random Assumption

**Benjamin M. Marlin**
Yahoo! Research and
Department of Computer Science
University of Toronto
Toronto, ON M5S 3H5

**Richard S. Zemel**
Department of
Computer Science
University of Toronto
Toronto, ON M5S 3H5

**Sam Roweis**
Department of
Computer Science
University of Toronto
Toronto, ON M5S 3H5

**Malcolm Slaney**
Yahoo! Research
Sunnyvale, CA 94089

## Abstract

Rating prediction is an important application, and a popular research topic in collaborative filtering. However, both the validity of learning algorithms, and the validity of standard testing procedures rest on the assumption that missing ratings are missing at random (MAR). In this paper we present the results of a user study in which we collect a random sample of ratings from current users of an online radio service. An analysis of the rating data collected in the study shows that the sample of random ratings has markedly different properties than ratings of user-selected songs. When asked to report on their own rating behaviour, a large number of users indicate they believe their opinion of a song *does* affect whether they choose to rate that song, a violation of the MAR condition. Finally, we present experimental results showing that incorporating an explicit model of the missing data mechanism can lead to significant improvements in prediction performance on the random sample of ratings.

## 1 Introduction

In a typical collaborative filtering system users assign ratings to items, and the system uses information from all users to recommend previously unseen items that each user might like or find useful. One approach to recommendation is to predict the ratings for all unrated items, and then recommend the items with the highest predicted ratings. Collaborative filtering research within the machine learning community has focused almost exclusively on developing new models and new learning procedures to improve rating prediction performance [2, 4, 5, 6, 8].

A critical assumption behind both learning methods and testing procedures is that the missing ratings are *missing at random* [7, p. 89]. One way to violate the missing at random condition in the collaborative filtering setting is for the probability of not observing a rating to depend on the value of that rating. In an internet-based movie recommendation system, for example, a user may be much more likely to see movies that they think they will like, and to enter ratings for movies that they see. This would create a systematic bias towards observing ratings with higher values.

Consider how this bias in the observed data impacts learning and prediction. In a nearest neighbour method it is still possible to accurately identify the neighbours of a given user [5]. However, the prediction for a particular item is based only on the available ratings of neighbours who rated the item in question. Conditioning on the set of users who rated the item can introduce bias into the predicted rating. The presence of non-random missing data can similarly introduce a systematic bias into the learned parameters of parametric and semi-parametric models including mixture models [1], customized probabilistic models [8], and matrix factorization models [2].

It is important to note that the presence of non-random missing data introduces a complementary bias into the standard testing procedure for rating prediction experiments [1] [5] [8, p.90]. Models are usually learned on one subset of the observed data, and tested on a different subset of the observed data. If the distribution of the observed data is different from the distribution of the fully completed data for any reason, the estimated error on the test data can be an arbitrarily poor estimate of the error on the fully completed data. Marlin, Roweis, and Zemel confirm this using experiments on synthetic data [9].

In this paper we present the results of the first study to analyze the impact of the missing at random assumption on collaborative filtering using data collected from real users. The study is based on users of Yahoo! Music's LaunchCast radio service. We begin with a review of the theory of missing data due to Little and Rubin [7]. We analyze the data that was gathered during the study, which includes survey responses, and ratings for *randomly* chosen songs. We describe models for learning and prediction with non-random missing data, and introduce a new experimental protocol for rating prediction based on training using user-selected items, and testing using randomly selected items. Experimental results show that incorporating a simple, explicit model of the missing data mechanism can lead to significant improvements in test error compared to treating the data as missing at random.

## 2 Missing Data Theory

A collaborative filtering data set can be thought of as a rectangular array $x$ where each row in the array represents a user, and each column in the array represents an item. $x_{im}$ denotes the rating of user $i$ for item $m$. Let $N$ be the number of users in the data set, $M$ be the number of items, and $V$ be the number of rating values. We introduce a companion matrix of response indicators $r$ where $r_{im} = 1$ if $x_{im}$ is observed, and $r_{im} = 0$ if $x_{im}$ is not observed. We denote any latent values associated with data case $i$ by $z_i$. The corresponding random variables are denoted with capital letters.

We adopt the factorization of the joint distribution of the data $X$, response indicators $R$, and latent variables $Z$ shown in Equation 2.1.

$$P(R, X, Z|\mu, \theta) = P(R|X, Z, \mu)P(X, Z|\theta) \quad (2.1)$$

We refer to $P(R|X, Z, \mu)$ as the missing data model or missing data mechanism, and $P(X, Z|\theta)$ as the data model. The intuition behind this factorization is that a complete data case is first generated according to the data model, and the missing data model is then used to select the elements of the data matrix that will not be observed.

### 2.1 Classification Of Missing Data

Little and Rubin classify missing data into several types including missing completely at random (MCAR), missing at random (MAR), and not missing at random (NMAR) [7, p. 14]. The MCAR condition is defined in Equation 2.2, and the MAR condition is defined in Equation 2.3. Under MCAR the response probability for an item or set of items cannot depend on the data values in any way. Under the MAR condition, the data vector is divided into a missing and an observed part according to the value of $r$ in question: $x = [x^{mis}, x^{obs}]$. The intuition is that the probability of observing a particular response pattern can only depend on the elements of the data vector that are observed under that pattern [10]. In addition, both MCAR and MAR require that the parameters $\mu$ and $\theta$ be distinct, and that they have independent priors.

$$P_{mcar}(R|X, Z, \mu) = P(R|\mu) \quad (2.2)$$
$$P_{mar}(R|X, Z, \mu) = P(R|X^{obs}, \mu) \quad (2.3)$$

Missing data is NMAR when the MAR condition fails to hold. The simplest reason for MAR to fail is that the probability of not observing a particular element of the data vector depends on the value of that element. In the collaborative filtering case this corresponds to the idea that the probability of observing the rating for a particular item depends on the user's rating for that item. When that rating is not observed, the missing data are not missing at random.

### 2.2 Impact Of Missing Data

When missing data is missing at random, maximum likelihood inference based on the observed data only is unbiased. We demonstrate this result in Equation 2.7. The key property of the MAR condition is that the response probabilities are independent of the missing data, allowing the complete data likelihood to be marginalized independently of the missing data model. However, when missing data is not missing at random, this important property fails to hold, and it is not possible to simplify the likelihood beyond Equation 2.4 [7, p. 219]. Ignoring the missing data mechanism will clearly lead to biased parameter estimates since an incorrect likelihood function is being used. For non-identifiable models such as mixtures, we will use the terms "biased" and "unbiased" in a more general sense to indicate whether the parameters are optimized with respect to the correct likelihood function.

$$\mathcal{L}_{mar}(\theta|x^{obs}, r)$$
$$= \int_{x^{mis}} \int_{z} P(X, Z|\theta) P(R|X, Z, \mu) dZ dX^{mis} \quad (2.4)$$
$$= P(R|X^{obs}, \mu) \int_{x^{mis}} \int_{z} P(X, Z|\theta) dZ dX^{mis} \quad (2.5)$$
$$= P(R|X^{obs}, \mu) P(X^{obs}|\theta) \quad (2.6)$$
$$\propto P(X^{obs}|\theta) \quad (2.7)$$

From a statistical perspective, biased parameter estimates are a serious problem. From a machine learning perspective, the problem is only serious if it adversely affects the end use of a particular model. Using synthetic data experiments, Marlin, Zemel, and Roweis

Table 1: User reported frequency of rating songs as a function of preference level.

| Rating Frequency | Preference Level | | | | |
|---|---|---|---|---|---|
| | Hate | Don't like | Neutral | Like | Love |
| Never | 6.76% | 4.69% | 2.33% | 0.11% | 0.07% |
| Very Infrequently | 1.59% | 4.17% | 9.46% | 0.54% | 0.35% |
| Infrequently | 1.63% | 4.44% | 24.87% | 1.48% | 0.20% |
| Often | 12.46% | 22.50% | 26.83% | 25.30% | 5.46% |
| Very Often | 77.56% | 64.20% | 36.50% | 72.57% | 93.91% |

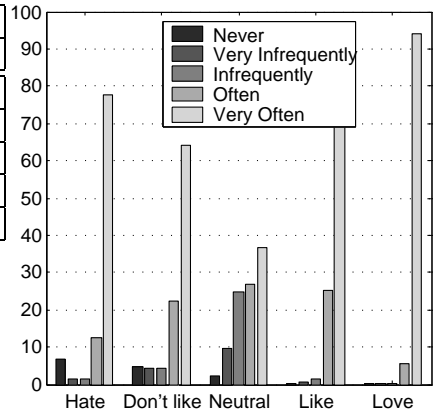

**Survey Results**: Yahoo! LaunchCast users were asked to report the frequency with which they choose to rate a song given their preference for that song. The data above show the distribution over rating frequencies given several preference levels. Users could select only one rating frequency per preference level.

demonstrated that ignoring the missing data mechanism in a rating prediction setting can have a significant impact on prediction performance [9].

## 3 Yahoo! LaunchCast Rating Study

To properly assess the impact of the missing at random assumption on rating prediction, we require a test set consisting of ratings that are a random sample of the ratings contained in the complete data matrix for a given set of users. In this section we describe a study conducted in conjunction with Yahoo! Music's LaunchCast Radio service to collect such a data set.

LaunchCast radio is a customizable streaming music service where users can influence the music played on their personal station by supplying ratings for songs. The LaunchCast Radio player interface allows the user enter a rating for the currently playing song using a five point scale. Users can also enter ratings for artists and albums. [1]

Data was collected from LaunchCast Radio users between August 22, 2006 and September 12, 2006. Users based in the US were able to join the study by clicking on a link in the LaunchCast player. Both the survey responses and rating data were collected through the study's web site. A total of 35,786 users contributed data to the study. Unless indicated otherwise, the results reported in this paper are based on a subset of 5400 survey participants who had at least 10 existing ratings in the LaunchCast rating database. The filtering we applied to the survey participants is required for the rating prediction experiments presented in Section 5.

---

[1] The Yahoo! Music LaunchCast web site is available at http://music.yahoo.com/launchcast/.

### 3.1 User Survey

The first part of the study consisted of a user survey containing sixteen multiple choice questions. The question relevant to this work asked users to report on how their preferences affect which songs they choose to rate. The question was broken down by asking users to estimate how often they rate a song given the degree to which they like it. The results are summarized in Table 1, and represented graphically in the accompanying figure. Each column in the table gives the results for a single survey question. For example, the column labeled "neutral" corresponds to the question "If I hear a song I feel neutral about I choose to rate it:" with the possible answers being "never", "very infrequently", "infrequently", "often", and "very often".

The results indicate that the choice to rate a song does depend on the user's opinion of that song. Most users tend to rate songs that they love more often than songs they feel neutral about, and somewhat more often than songs that they hate. Users were also directly asked if they thought their preferences for a song *do not* affect whether they choose to rate it. 64.85% of users responded that their preferences *do* affect their choice to rate a song. By contrast, the missing at random assumption requires that the underlying ratings not influence a users choice to rate a song.

### 3.2 Rating Data Collection

Following the survey, users were presented with a set of ten songs to rate. The artist name and song title were given for each song, along with a thirty second audio clip, which the user could play before entering a rating. Ratings were entered on the standard five point scale used by Yahoo! Music. The set of ten songs presented to each user was chosen at random without replacement from a fixed set of 1000 songs. The fixed

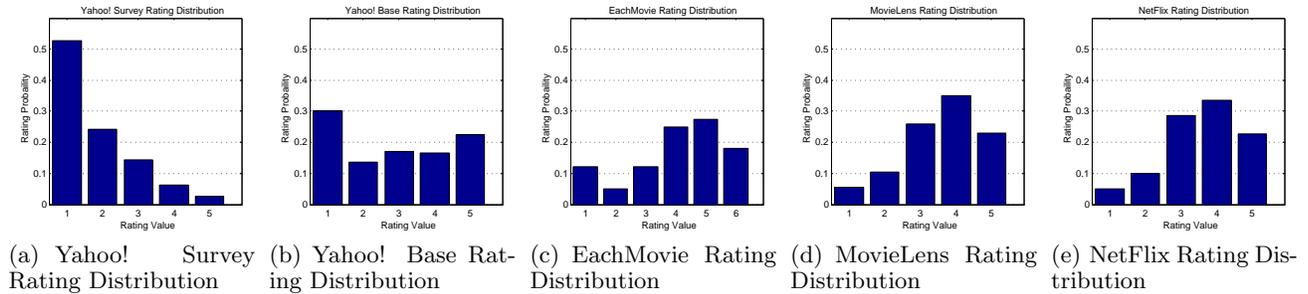

(a) Yahoo! Survey Rating Distribution  (b) Yahoo! Base Rating Distribution  (c) EachMovie Rating Distribution  (d) MovieLens Rating Distribution  (e) NetFlix Rating Distribution

Figure 1: Distribution of rating values in the Yahoo! survey set and base set compared to several popular collaborative filtering data sets including EachMovie, MovieLens, and Netflix.

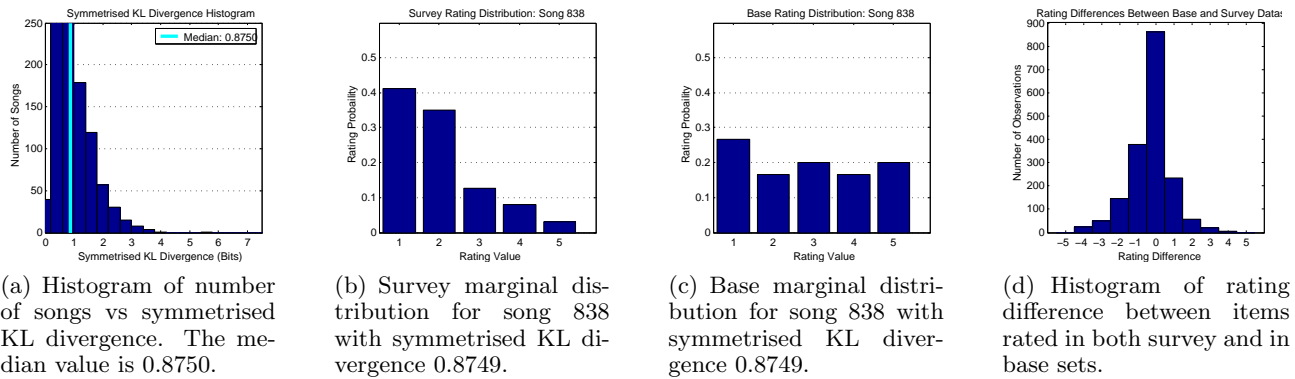

(a) Histogram of number of songs vs symmetrised KL divergence. The median value is 0.8750.  (b) Survey marginal distribution for song 838 with symmetrised KL divergence 0.8749.  (c) Base marginal distribution for song 838 with symmetrised KL divergence 0.8749.  (d) Histogram of rating difference between items rated in both survey and in base sets.

Figure 2: Panels (a) to (c) give an indication of the distribution of differences between survey and base marginal distributions for each song. Panel (d) shows the histogram of differences between ratings for songs that were observed in both the survey, and the LaunchCast database. This histogram was computed based on all 35, 786 survey participants.

set of 1000 songs used in the survey were chosen at random from all the songs in the LaunchCast play list having at least 500 existing ratings in the LaunchCast rating database.

We refer to ratings collected during the survey as "survey ratings." In addition, each survey participant's existing ratings on the set of 1000 survey songs was extracted from the LaunchCast database. We refer to these existing ratings as the "base ratings." The survey ratings represent a random sample of songs for each survey participant, while the base ratings represent ratings for songs the survey participant *chose* to enter. We repeat that unless otherwise indicated, the results we report are restricted to the subset of 5400 survey participants with at least 10 base ratings.

Figures 1(a) and 1(b) show the empirical distribution of survey ratings and base ratings for the 5400 survey participants. These figures show a dramatic difference between the two distributions. The number of four and five star rating values is many times lower in the survey set than the base set. The difference between the survey and base distributions is not surprising given that users can influence the LaunchCast system to play songs reflecting their preferences. Figures 1(c) to 1(e) give the rating distributions for several other collaborative filtering data sets including EachMovie, MovieLens, and Netflix. All these distributions show a much higher proportion of high rating values than are present in the random sample we collected during the survey.

To further analyze the difference between the base ratings and the survey ratings, we computed the distribution over rating values for each item. For a particular item $m$ let $P^S(X_m = v)$ be the empirical probability of rating value $v$ in the survey set, and $P^B(X_m = v)$ be the empirical probability of rating value $v$ in the base set. We smooth the empirical probabilities by one count per rating value to avoid zeros. We use the symmetrised Kullback–Leibler divergence (SKL) shown in Equation 3.8 to measure the difference between the $P^S(X_m = v)$ and $P^B(X_m = v)$ distributions for each item $m$.

$$SKL_m = \sum_{v=1}^{V} P^S(X_m = v) \log\left(\frac{P^S(X_m = v)}{P^B(X_m = v)}\right)$$
$$+ P^B(X_m = v) \log\left(\frac{P^B(X_m = v)}{P^S(X_m = v)}\right) \quad (3.8)$$

Figure 2(a) shows a histogram of the symmetrised Kullback−Leibler divergence values. The thick vertical line in the plot indicates the median SKL value of 0.8750 bits. Song 838 has an SKL value of 0.8749 bits, the largest SKL value less than the median. Figures 2(b) and 2(c) illustrate the marginal rating distributions for song 838. These distributions are qualitatively quite different, and half of the songs in the survey set exhibit a more extreme difference according to the SKL measure.

A pertinent question is whether users' ratings reported during the survey were consistent with ratings recorded during normal use of the LaunchCast system. To help answer this question we extracted the set of ratings that were observed both in the survey, and in the LaunchCast data set. Figure 2(d) shows a histogram of the differences $x_{im}^B - x_{im}^S$ where the user-song pair $(i, m)$ is observed in both the survey $S$ and base sets $B$. We can see from Figure 2(d) that the agreement between the two sets of ratings is quite good. Note that this comparison is based on the complete set of 35,786 survey participants. The intersection of the survey and base sets contained approximately 1700 observations.

It is important to note that the observed discrepancy between the survey set marginal distributions and the base set marginal distributions is not conclusive evidence that the missing data in the base set is NMAR. This is due to the fact that the MAR assumption can hold for the true underlying data model, but not for more simplistic models like the marginal model used in the present analysis. Nevertheless, we believe that the results of this analysis combined with the results of the user survey provide compelling evidence against the MAR assumption.

## 4 Modeling Non-Random Missing Data

Many probabilistic models have the property that missing data can be analytically integrated away under the missing at random assumption. This allows for computationally efficient, unbiased parameter estimation. The multinomial mixture model has this convenient property, and it has been well studied in the collaborative filtering domain [8].

**Algorithm 1** MAP EM Algorithm for the Bayesian multinomial mixture model.

**E-Step:**
$$q_{zi} \leftarrow \frac{\theta_z \prod_{m=1}^{M} \prod_{v=1}^{V} \beta_{vmz}^{r_{im}[x_{im}=v]}}{\sum_{z=1}^{K} \theta_z \prod_{m=1}^{M} \prod_{v=1}^{V} \beta_{vmz}^{r_{im}[x_{im}=v]}}$$

**M-Step:**
$$\theta_z \leftarrow \frac{\alpha_z - 1 + \sum_{i=1}^{N} q_{zi}}{\sum_{z=1}^{K}(\alpha_z + \sum_{i=1}^{N} q_{zi}) - K}$$
$$\beta_{vmz} \leftarrow \frac{\phi_{vmz} - 1 + \sum_{i=1}^{N} q_{zi} r_{im}[x_{im}=v]}{\sum_{v=1}^{V} \phi_{vmz} - V + \sum_{i=1}^{N} q_{zi} r_{im}}$$

When the missing at random assumption is not believed to hold, Equation 2.4 shows that parameter estimation will be biased unless the true missing data mechanism is known. In a domain as complex and high dimensional as collaborative filtering, a more realistic goal is to formulate models of the missing data mechanism that capture some of its key properties.

In this section we present the basic multinomial mixture model, and give learning and prediction methods under the MAR assumption. We extend the mixture model by combining it with a Bayesian variant of the *CPT-v* missing data model [9], which captures a key property of the non-random missing data mechanism implied by the user survey results. We give learning and prediction methods for the combined mixture/*CPT-v* model.

### 4.1 Multinomial Mixture Data Model

The multinomial mixture model is a generative probabilistic model. It captures the simple intuition that users form groups or clusters according to their preferences for items. We summarize the probabilistic model below.

$$P(\theta, \beta | \alpha, \phi) = D(\theta|\alpha) \prod_z \prod_m D(\beta_{mz}|\phi_{mz}) \quad (4.9)$$

$$P(Z_i = z|\theta) = \theta_z \quad (4.10)$$

$$P(\boldsymbol{X}_i = \boldsymbol{x}_i | Z_i = z, \beta) = \prod_m \prod_v \beta_{vmz}^{[x_{im}=v]} \quad (4.11)$$

The main feature of the model is the variable $Z_i$, which indicates which of the $K$ groups or clusters user $i$ belongs to. To generate a complete data vector $\boldsymbol{X}_i$ for user $i$, a value $k$ for $Z_i$ is first sampled according to the discrete distribution $P(Z_i = z|\theta)$. A rating value $v$ for each item $m$ is then sampled independently from the discrete distribution $P(X_{im} = v|Z_i = z, \boldsymbol{\beta}_{mk})$. Importantly, all we observe is the final data vector $\boldsymbol{X}_i$. $Z_i$ is considered a latent variable since its value is never observed.

In a Bayesian mixture model, the parameters $\theta$ and $\boldsymbol{\beta}_{mz}$ are also regarded as random variables. Before generating any data cases, the model parameters are first sampled from their prior distributions. The same model parameters are assumed to generate all data cases. In the present case we choose conjugate Dirichlet priors for both $\theta$, and $\boldsymbol{\beta}_{mz}$. We give the form of the Dirichlet priors for $\theta$ and $\boldsymbol{\beta}_{mz}$ in Equations 4.12 and 4.13.

$$D(\theta|\alpha) = \frac{\Gamma(\sum_{z=1}^{K}\alpha_k)}{\prod_{z=1}^{K}\Gamma(\alpha_z)}\prod_{z=1}^{K}\theta_z^{\alpha_z-1} \quad (4.12)$$

$$D(\beta_{mk}|\phi_{mz}) = \frac{\Gamma(\sum_{v=1}^{V}\phi_{vmz})}{\prod_{v=1}^{V}\Gamma(\phi_{vmz})}\prod_{v=1}^{V}\beta_{vmz}^{\phi_{vmz}-1} \quad (4.13)$$

The posterior log probability of the mixture model parameters $\theta$ and $\boldsymbol{\beta}_{mk}$ given a sample of incomplete data is shown below in Equation 4.14.

$$\mathcal{L}_{mar} = \sum_{i=1}^{N}\log\left(\sum_{z=1}^{K}\theta_z\prod_{m=1}^{M}\prod_{v=1}^{V}\beta_{vmz}^{r_{im}[x_{im}=v]}\right)$$
$$+ \log D(\theta|\alpha) + \sum_{m=1}^{M}\sum_{z=1}^{z}\log D(\beta_{mz}|\phi_{mz}) \quad (4.14)$$

The Bayesian mixture model parameters are learned from incomplete data by maximizing the posterior log probability of the observed data. This optimization is efficiently performed using the Expectation Maximization (EM) algorithm of Dempster, Laird, and Rubin [3]. We give the maximum a posteriori (MAP) EM algorithm for the Bayesian multinomial mixture model in Algorithm 1. In the expectation step of the algorithm we compute posterior distribution on $Z_i$ for each user $i$ given the current values of the model parameters. This inference procedure is also important for prediction. We give it in Equation 4.15.

$$P(Z_i = z|\boldsymbol{x}_i, \boldsymbol{r_i}, \theta, \beta) = \frac{\theta_z\prod_{m=1}^{M}\prod_{v=1}^{V}\beta_{vmz}^{r_{im}[x_{im}=v]}}{\sum_{z=1}^{K}\theta_z\prod_{m=1}^{M}\prod_{v=1}^{V}\beta_{vmz}^{r_{im}[x_{im}=v]}} \quad (4.15)$$

### 4.2 The *CPT-v* Missing Data Model

The *CPT-v* missing data model was proposed by Marlin, Roweis, and Zemel as one of the simpler non-random missing data model [9]. The *CPT-v* model captures the intuition that a user's preference for an item affects whether they choose to rate that item or not. The model assumes that the choice to rate each

**Algorithm 2** MAP EM Algorithm for the Bayesian multinomial mixture/*CPT-v* model.

**E-Step:**
$\lambda_{vmzn} \leftarrow ([x_{im}=v]\mu_v\beta_{vmz})^{r_{im}}((1-\mu_v)\beta_{vmz})^{1-r_{im}}$
$\gamma_{mzn} \leftarrow \sum_{v=1}^{V}\lambda_{vmzn}$
$q_{zi} \leftarrow \frac{\theta_{z_n}\prod_{m=1}^{M}\gamma_{mzn}}{\sum_{z=1}^{K}\theta_{z'}\prod_{m=1}^{M}\gamma_{mzn}}$

**M-Step:**
$\theta_z \leftarrow \frac{\alpha_z-1+\sum_{i=1}^{N}q_{zi}}{\sum_{z=1}^{K}(\alpha_z+\sum_{i=1}^{N}q_{zi})-K}$
$\beta_{vmz} \leftarrow \frac{\phi_{vmk}-1+\sum_{i=1}^{N}\phi_{zi}\lambda_{vmzn}/\gamma_{mzn}}{\sum_{v=1}^{V}\phi_{vmk}-V+\sum_{n=1}^{N}q_{zi}}$
$\mu_v \leftarrow \frac{\xi_{1v}-1+\sum_{i=1}^{N}\sum_{z=1}^{K}q_{zi}\sum_{m=1}^{M}r_{mn}\lambda_{vmzn}/\gamma_{mzn}}{\xi_{0v}+\xi_{1v}-2+\sum_{n=1}^{N}\sum_{z=1}^{K}q_{zi}\sum_{m=1}^{M}\lambda_{vmzn}/\gamma_{mzn}}$

item is independent, and that the probability of rating a single item, given that the user's rating for that item is $v$, is Bernoulli distributed with parameter $\mu_v$. We extend the basic *CPT-v* model slightly by introducing a Beta prior on the parameters $\mu_v$. The probabilistic model is summarized below.

$$P(\mu|\xi) = \prod_v Beta(\mu_v|\xi_v) \quad (4.16)$$

$$P(\boldsymbol{R}=\boldsymbol{r}|\boldsymbol{X}=\boldsymbol{x}) = \quad (4.17)$$
$$\prod_{m=1}^{M}\prod_{v=1}^{V}\mu_v^{r_{im}[x_{im}=v]}(1-\mu_v)^{(1-r_{im})[x_{im}=v]}$$

The Beta prior we select is the conjugate prior for the Bernoulli parameters $\mu_v$. We give the form of the prior distribution in Equation 4.18.

$$B(\mu_v|\xi_v) = \frac{\Gamma(\xi_{0v}+\xi_{1v})}{\Gamma(\xi_{0v})\Gamma(\xi_{1v})}\mu_v^{\xi_{1v}-1}(1-\mu_v)^{\xi_{0v}-1} \quad (4.18)$$

The factorized structure of the model is quite restrictive. However, it allows the missing data to be summed out of the posterior distribution leaving local factors that only depend on one missing data value at a time. The log posterior distribution on the model parameters is given in Equation 4.19.

$$\mathcal{L}_{CPTv} = \sum_{n=1}^{N}\log\left(\sum_{z=1}^{K}\theta_z\prod_{m=1}^{M}\gamma_{mzn}\right) + \sum_{v=1}^{V}\log B(\mu_v|\xi_v) \quad (4.19)$$

$$\gamma_{mzn} = \begin{cases} \prod_v(\mu_v\beta_{vmz})^{[x_{im}=v]} & \dots r_{im}=1 \\ \sum_v(1-\mu_v)\beta_{vmz} & \dots r_{im}=0 \end{cases}$$

As in the standard Bayesian mixture model case, the log posterior distribution of the combined Bayesian mixture/*CPT-v* model can be optimized using an expectation maximization algorithm. We give the details in Algorithm 2. Again, inference for the latent mixture indicator $Z_i$ is the main operation in the expectation step. As we can see in Equation 4.20, the form of the inference equation is very similar to the standard mixture case.

$$P(Z_i = z | \boldsymbol{x_i}, \boldsymbol{r_i}, \theta, \beta) = \frac{\theta_z \prod_{m=1}^{M} \gamma_{mzn}}{\sum_{z=1}^{K} \theta_z \prod_{m=1}^{M} \gamma_{mzn}} \quad (4.20)$$

### 4.3 Rating Prediction

To make a prediction for user $i$ and item $m$ we first need to perform inference in the model to compute the posterior distribution $P(Z_i = z | \boldsymbol{x_i}, \boldsymbol{r_i}, \theta, \beta)$ over the mixture indicator variable $Z_i$. For the multinomial mixture model under the MAR assumption we use Equation 4.15. For the multinomial mixture model combined with the *CPT-v* model we use Equation 4.20. For both models, we compute the predictive distribution over rating values for item $m$ according to Equation 4.21.

$$P(X_{im} = v) = \sum_{z=1}^{K} \beta_{vmz} P(Z_i = z | \boldsymbol{x_i}, \boldsymbol{r_i}, \theta, \beta) \quad (4.21)$$

## 5 Experimental Method and Results

Both the analysis of the user survey, and the analysis of the rating data collected in this study suggest that missing data in the LaunchCast database is not missing at random. The question we address in this section is whether treating the missing data as if it were not missing at random leads to an improvement in predictive performance relative to treating the missing data as if it were missing at random. We discuss the data set used for rating prediction experiments, the methods tested, the experimental protocol, and the results.

### 5.1 Rating Prediction Data Set

The rating data set used in the experiments is based on the 1000 survey songs and 10,000 users. The set of 10,000 users consists of a random selection of 5000 of the 5400 survey participants with at least 10 existing ratings on the set of 1000 survey songs, and a random selection of 5000 non-survey users with at least 10 existing ratings on the set of 1000 survey songs. We chose to enforce a minimum number of existing ratings per user so that rating prediction methods would have at least 10 observations on which to base predictions. Non-survey users were included to provide more training data to the learning methods. The 400 held out survey participants will later be used for additional parameter estimation tasks.

The rating data is divided into a test set consisting of the survey ratings collected for the 5000 survey participants, and a training set consisting of the existing ratings extracted from the LaunchCast database for each of the 10,000 users. Thus, the test set contains 10 ratings for 10 songs chosen completely at random from the set of 1000 survey songs for each of the 5000 survey participants, giving a total of 50,000 ratings. The training set consists of a minimum of 10 ratings for all users giving a total of approximately 218,000 ratings. The ratings in the training set are ratings for the 1000 survey songs entered by users during normal use of the LaunchCast music service. Any overlapping ratings in the training and test sets were removed from the training set before selecting users for the data set.

### 5.2 Rating Prediction Experiments

The experimental protocol we follow is to train the models on the training set, and test on the test set defined in the previous section. The novel aspect of the protocol stems from the division of the rating data into a test set consisting of a random sample of ratings for each user, and a training set consisting of a possibly non-random sample of ratings for each user.

The baseline method for the rating prediction experiments is the Bayesian multinomial mixture model under the MAR assumption. We learn the model parameters using the MAP-EM algorithm given in Algorithm 1 with the prior parameters $\phi_{vmz} = 2$ and $\alpha_z = 2$ for all $v, m, z$. We run EM until either the log posterior attains a relative convergence of $10^{-5}$, or 1000 EM iterations are performed.

We compare the baseline multinomial mixture model, which ignores the missing data mechanism (denoted *MM/None*), to a combination of the Bayesian multinomial mixture with the *CPT-v* missing data model (denoted *MM/CPT-v*). We learn the model parameters using the MAP-EM algorithm given in Algorithm 2 with the prior parameters $\phi_{vmz} = 2$ and $\alpha_z = 2$ for all $v, m, z$. We run EM until either the log posterior attains a relative convergence of $10^{-5}$, or 1000 EM iterations are performed.

We use the learned models to predict the value of each test and training rating for each user. We report prediction error in terms of mean absolute error (MAE) [1]. Specifically, we use Equation 4.21 to compute the posterior predictive distribution for a given song. We then predict the median value of the posterior predictive distribution.

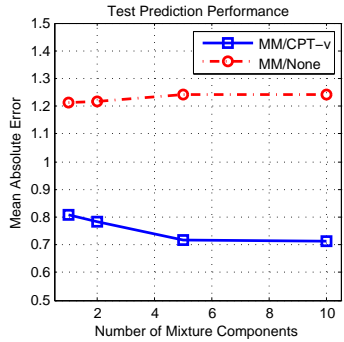
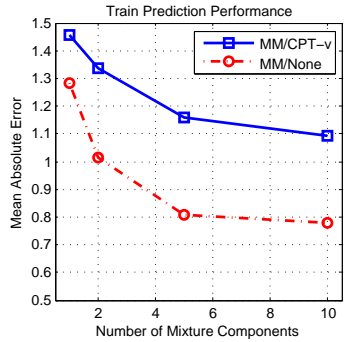
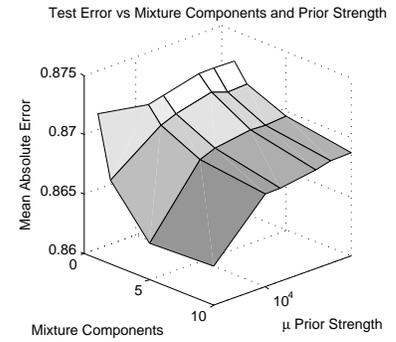

(a) Best case test set prediction error for MM/CPT-v vs MM/None.

(b) Best case training set prediction error for MM/CPT-v vs MM/None.

(c) MM/CPT-v test set prediction error vs prior strength and number of mixture components.

Figure 3: Panel 3(a) shows the test error for the multinomial mixture/*CPT-v* model (MM/CPT-v) using best-case parameters for the missing data model compared to the standard multinomial mixture model with no missing data model (MM/None model). Panel 3(b) shows the same comparison for training error. Panel 3(c) shows a plot of the test error for MM/CPT-v as the number of mixture components and the prior strength vary.

### 5.3 Rating Prediction Results

The main question we are interested in is how much of a gain in rating prediction performance can be obtained by treating the missing data as if it were non-random? To get a sense of the best case performance of the *CPT-v* model we estimated an optimal set of $\mu$ parameters using held out survey ratings. Under the simplified data model defined by $P(X_{im} = v) = \beta_v$, $\mu_v$ can be directly estimated as:

$$\mu_v = \delta_v/\beta_v \qquad (5.22)$$

where $\delta_v = P(R_{im} = 1, X_{im} = v)$. We use survey ratings from the 400 survey users not included in the prediction data set to estimate the parameters $\beta_v$. This is a valid estimate for $\beta_v$ since the missing survey ratings are missing completely at random. We use previously existing ratings for the same 400 survey users to estimate the parameters $\delta_v$. This is a valid estimate for $\delta_v$ since the missing ratings in the LaunchCast database are subject to the missing data mechanism that we wish to model. The set of missing data model parameters estimated using this method is $\hat{\mu} = [0.014, 0.011, 0.027, 0.063, 0.225]$. Recall that $\mu_v$ is is the probability of observing a rating given that its value is $v$.

We fixed $\mu$ to these values and ran the EM algorithm given in Algorithm 2 to estimate the mixture model parameters only. We then computed the prediction error on the testing and training sets. This experiment was performed using 1, 2, 5, and 10 mixture components. Five repetitions were performed for each number of mixture components, and the results were averaged. The multinomial mixture model with no missing data model was learned using Algorithm 1, and tested using exactly the same procedure.

Figure 3(a) gives a comparison of the average test error obtained by the combined multinomial mixture and *CPT-v* model (MM/CPT-v), and that obtained by the multinomial mixture model with no missing data model (MM/None). The best average test error obtained by MM/CPT-v is 0.7148 using ten mixture components, while the best average test error obtained by MM/None is 1.2126 using one mixture component. MM/CPT-v obtains a reduction in test error of over 40% relative to MM/None. Note that the standard error of the mean is less than 0.01 for both models. This clearly shows that there is a large benefit to treating the missing data as if it *were not* missing at random.

It is interesting to observe that the test error actually increases slightly for MM/None as the number of mixture components increases. Increasing the complexity of the model allows it to match the distribution of training data more closely as seen in Figure 3(b), to the detriment of test performance.

We performed a second set of experiments where both the mixture model and missing data model parameters of the MM/CPT-v model were learned. An informative prior for the $\mu$ parameters was defined using $\hat{\mu}$ as $\xi_{1v} = S\hat{\mu}_v$, and $\xi_{0v} = S(1 - \hat{\mu}_v)$. $S$ is the prior strength. We tested between 1 and 10 mixture components, and prior strengths between 200 and 100,000. Five repetitions of the experiment were performed for each combination of prior strength and number of mixture components.

Figure 3(c) shows the mean test error for each com-bination. The maximum standard error of the mean is less than 0.001. The results show that even at relatively low values of the prior strength, the MDM/CPT-v model obtains a significant improvement in test error over the baseline mixture model. However, even at seemingly large values of the prior strength, the performance does not approach that of MM/CPT-v with $\mu$ fixed to $\hat{\mu}$. The small range in test values found in this experiment appears to result from the posterior having a very strong mode at a solution where almost all of the missing data is explained as having the value 2. This type of boundary solution was previously observed for the maximum likelihood version of the MM/CPT-v model [9].

## 6 Discussion and Conclusions

In the collaborative filtering domain, both the validity of learning algorithms, and the validity of standard testing procedures rests on the assumption that missing rating data is missing at random. In this paper we have provided compelling evidence of a violation of the missing at random assumption in real collaborative filtering data. Furthermore, we have shown that incorporating an explicit model of the missing data mechanism can significantly improve rating prediction on a test set.

Results of the LaunchCast user survey indicate that users are aware that their preferences impact which items they choose to rate. Ratings of randomly selected songs collected in this study show systematic differences relative to ratings of user selected songs. We introduced a new experimental protocol where models are trained on ratings of user selected songs, and tested on ratings of randomly selected songs. Using this protocol we found that the *CPT-v* missing data model leads to a surprising boost in test performance relative to ignoring the missing data mechanism, if a suitable set of missing data parameters can be learned or estimated.

We have shown that a relatively small number of ratings for songs chosen at random can be used to estimate a set of missing data model parameters that generalizes very well to a larger population of users. The main shortcoming of this work is that even given strong prior information, learning missing data model parameters still results in solutions of significantly lower quality than when the missing data model parameters are estimated using held out ratings. The use of Markov Chain Monte Carlo inference may lead to better predictive performance than the current MAP framework if the boundary solutions found by MAP EM have little posterior mass. The use of MCMC methods would also allow us to consider more flexible data and missing data models including hierarchical, and non-parametric constructions.

## Acknowledgements

This research was supported by the Natural Sciences and Engineering Research Council of Canada (NSERC).